\documentclass[letterpaper]{article}
\usepackage{aaai20}
\usepackage{times}
\usepackage{helvet}
\usepackage{courier}
\usepackage[hyphens]{url} 
\usepackage{graphicx}
\usepackage{graphicx}
\usepackage{amsmath}
\usepackage{mathtools}
\usepackage{bbm}
\usepackage[linesnumbered,ruled,vlined]{algorithm2e}

\title{CAG: A Real-time Low-cost Enhanced-robustness High-transferability Content-aware Adversarial Attack Generator}
\author{Huy Phan$^{1*}$, Yi Xie$^1$\protect\thanks{Equal contribution}, Siyu Liao$^1$, Jie Chen$^2$, Bo Yuan$^1$ \\
$^1$Rutgers University - School of Engineering \\
$^2$MIT-IBM Watson AI Lab, IBM Research\\
\{huy.phan, yi.xie, siyu.liao\}@rutgers.edu, chenjie@us.ibm.com, bo.yuan@soe.rutger.edu}

\begin{document}
\maketitle
\begin{abstract}
Deep neural networks (DNNs) are vulnerable to adversarial attack despite their tremendous success in many artificial intelligence fields. Adversarial attack is a method that causes the intended misclassfication by adding imperceptible perturbations to legitimate inputs. To date, researchers have developed numerous types of adversarial attack methods. However, from the perspective of practical deployment, these methods suffer from several drawbacks such as long attack generating time, high memory cost, insufficient robustness and low transferability. To address the drawbacks, we propose a Content-aware Adversarial Attack Generator (CAG) to achieve real-time, low-cost, enhanced-robustness and high-transferability adversarial attack. First, as a type of generative model-based attack, CAG shows significant speedup (at least 500 times) in generating adversarial examples compared to the state-of-the-art attacks such as PGD and C\&W. Furthermore, CAG only needs a single generative model to perform targeted attack to any targeted class. Because CAG encodes the label information into a trainable embedding layer, it differs from prior generative model-based adversarial attacks that use $n$ different copies of generative models for $n$ different targeted classes. As a result, CAG significantly reduces the required memory cost for generating adversarial examples. Moreover, CAG can generate adversarial perturbations that focus on the critical areas of input by integrating the class activation maps information in the training process, and hence improve the robustness of CAG attack against the state-of-art adversarial defenses. In addition, CAG exhibits high transferability across different DNN classifier models in black-box attack scenario by introducing random dropout in the process of generating perturbations. Extensive experiments on different datasets and DNN models have verified the real-time, low-cost, enhanced-robustness, and high-transferability benefits of CAG.
\end{abstract}

\section{Introduction}
Deep neural networks (DNNs) have achieved unprecedented success in many artificial intelligence fields, such as computer vision, natural language processing and speech recognition \cite{he2016deep,conneau2016very,yu2016automatic}. Despite their current popularity and prosperity, DNNs are still facing several severe challenges, especially their high vulnerability to adversarial attack \cite{goodfellow2014explaining,carlini2017towards}, which adds well-designed tiny perturbations to the legitimate inputs to cause the intended misclassification of DNN models. Such attacks could cause severe safety, economic and social problems if launched to the DNNs deployed in practical applications ranging from face recognition, autonomous driving to speech authentication.

\begin{figure}[ht]
\centering
\includegraphics[width=0.47\textwidth]{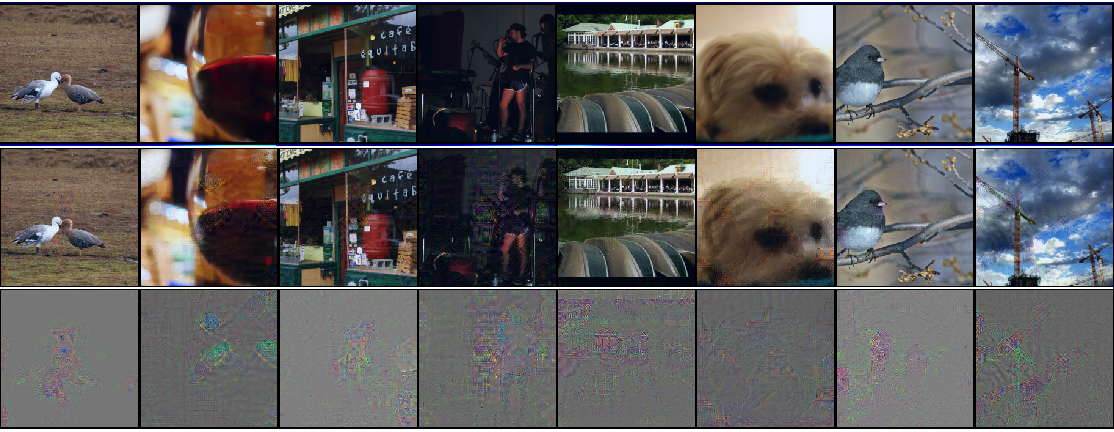}
\caption{Adversarial images generated with CAG using ImageNet dataset. From top row to bottom row: legitimate images, adversarial images, perturbations (enhanced).}
\label{adversarial_images}
\end{figure}

In order to address this critical challenge, the machine learning community has conducted extensive researches on the vulnerability of DNNs, from both the attack and defense aspects. Adversarial attack technique was pioneered by Szegedy \textit{et al.} \shortcite{szegedy2013intriguing}. Since then, researchers have developed various adversarial attacking algorithms, targeting different types of DNN models including convolutional neural networks, recurrent neural networks and graph neural networks, and also different application scenarios, ranging from image classification, machine translation, to graph classification etc. Among those algorithms, one popular class of attack techniques is fast gradient sign method (FGSM), which performs one-step gradient computation to craft untargeted adversarial examples \cite{goodfellow2014explaining}. Considering the relatively weak attack performance of FGSM, the machine learning community has proposed several iterative optimization-based techniques including C\&W, I-FGSM and PGD that deliver the state-of-the-art attack performance \cite{carlini2017towards,goodfellow2014explaining,madry2017towards}. Furthermore, some recent work has also proposed to use generative models, e.g., GAN and U-Net, to generate adversarial examples \cite{poursaeed2018generative,xiao2018generating,goodfellow2014generative,ronneberger2015u}.

Although the existing adversarial attack methods can already exhibit high attack success rate (ASR), especially in white-box attack scenario, from the perspective of practical deployment, they are still suffering one or more drawbacks, namely long adversarial example generating time, high memory cost for launching adversarial attack, insufficient robustness against defense methods and low transferability in black-box attack scenario.

Aiming to overcome these drawbacks, in this paper we propose a Content-aware Adversarial Attack Generator (CAG), to achieve real-time, low-cost, enhanced-robustness and high-transferability adversarial attack. We show some adversarial images generated by CAG in Figure \ref{adversarial_images}. The features and benefits of CAG are summarized as follows:

\begin{itemize}
    \item CAG is a generative model-based attack, so it can avoid time-consuming iterative optimization procedure to generate adversarial examples. Compared with the state-of-the-art iterative attacks such as PGD and C\&W, CAG achieves significant speedup (at least $500$ times), and hence makes real-time attack possible.
    \item CAG utilizes a trainable embedding layer to encode all label information to one single model, unlike prior generative model-based methods which require different generative models for different targeted classes. In $n$-class targeted attack scenario, the number of the required generative models is reduced from $n$ to $1$, thereby drastically reducing the memory cost for launching attacks.
    \item CAG integrates the class activation maps (CAMs) information into the training process, in contrast to many other attack methods that generate adversarial perturbations over the entire input. Consequently, CAG is able to generate adversarial perturbations that focus on the critical areas of input, and thus improves the attack's robustness against the state-of-art defense approaches.
    \item CAG exhibits high transferability across different DNN classifier models in black-box attack scenario. CAG can generate adversarial perturbations with better generality by introducing random dropout in the perturbations-generation process. As a result, CAG's adversarial examples have higher transferability when attacking unseen classifiers.
\end{itemize}

The rest of this paper is organized as follows. Section 2 introduces the related work on adversarial attack and defense methods. Section 3 discusses our motivation. Section 4 describes the technical details of CAG. The experimental results are presented and analyzed in Section 5. Section 6 draws the conclusions of all findings in our paper.

\section{Related Work}

\subsection{Adversarial Attacks}
To define an adversarial attack, let $X=\{x_1,...,x_m\}$ be a set of the valid inputs from the dataset, $y\in \{1, ..., L\}$ be the valid class label, and $F(\cdot)$ be the well-trained DNN classifier. Let $(x_{i}, y_{i})$ denote the $i$-th benign instance and the corresponding true label. The goal of an adversarial attack is to create the $x_{i}'=x_{i}+\delta$, where $\delta $ is imperceptible adversarial perturbation. A nontargeted attack requires $F(x_{i})\neq y_{i} $ and a targeted attack specifies $t\neq i$ such that $F(x_{i}')= y_{t}$.

\subsubsection{FGSM} Fast gradient sign method (FGSM) is a one-step fast-adversarial-example-generation approach \cite{goodfellow2014explaining}. It aims to linearize loss function in $L_\infty$ neighborhood of a legitimate input and to find the exact maximum of the linearized loss function. Correspondingly, its adversarial example generation formula is as follows:
\begin{equation*}
    x' = x + \epsilon \cdot sign(\Delta J(x,y_{true})),
\end{equation*}
where $y_{true}$ denotes the true label, $\Delta J(.,.)$ computes the gradient of the loss function, and $sign$ denotes the sign function. Notice that here $\epsilon$ is the attack strength parameter to control the balance between the attack performance and the norm of the perturbations.

\subsubsection{I-FGSM \& PGD} Although FGSM is fast, its attack performance is relatively weak. Researchers have proposed various approaches to achieve stronger attack by improving the vanilla FGSM method. Kurakin \textit{et al.} \shortcite{kurakin2016adversarial} propose to take multiple steps of FGSM (I-FGSM) with smaller attack strength $\alpha$ in an iterative way:
\begin{equation*}
    x'_{N+1} = Clip_{\epsilon} \{x'_N + \alpha \cdot sign(\Delta J(x'_N, y_{true})\},
\end{equation*}
where $x'_N$ is the adversarial image at the $N$-th iteration, and $Clip\{\cdot\}$ clips the overall attack strength back to $\epsilon$ at the end of the iteration. Notice that in the case of using $L_{\infty}$ norm, I-FGSM is equivalent to another popular iteration-based attack method (PGD) \cite{madry2017towards}.

\subsubsection{C\&W}
C\&W \cite{carlini2017towards} is an optimization-based attack method. It aims to optimize the loss function as follows:
\begin{equation*}
\begin{multlined}
\Vert x'-x\Vert_p + c\cdot \max(\underset{i \neq t}{\max} f(x')_i - f(x')_t, -\kappa),
\end{multlined}
\end{equation*}
where $t$ is the targeted class, $f(\cdot)$ denotes the softmax function, $c$ is a constant set by binary search, and $\kappa$ is an adjustable parameter that encourages the attacker to find an adversarial example being classified as class $t$ with high confidence. By minimizing the above loss function using Adam optimizer in an iterative way, C\&W can achieve high ASR with low perturbation norm.

\subsubsection{Generative Model-based} One drawback of the iterative methods mentioned above is long generating time. Hence another method to generate adversarial examples is to use a generative model, such as GAN, Autoencoder \cite{goodfellow2014generative} or U-Net. For instance, Xiao \textit{et al.} \shortcite{xiao2018generating} apply AdvGAN to craft perceptually realistic adversarial examples. Moreover, Baluja \textit{et al.}  \shortcite{baluja2017adversarial} develop an adversarial transformation network to convert inputs into adversarial examples. Poursaeed \textit{et al.} \shortcite{poursaeed2018generative} propose a method they name Generative Adversarial Perturbations (GAP) that uses a ResNet-based generative model \cite{johnson2016perceptual} to perform adversarial attack.

\subsection{Adversarial Defenses}

\subsubsection{Pixel Deflection}
The key idea of pixel deflection defense is to randomly replace pixels with nearby pixels \cite{prakash2018deflecting}. To achieve the replacement, this method uses CAMs of the top-5 predictions to guide the update of the pixels \cite{zhou2016learning}. In this scenario, the probability of a pixel being updated is inversely proportional to the likelihood that the area contains the object. After the pixel replacement, a denoising operation is applied to recover the classification accuracy.

\subsubsection{Randomization}
The mitigation of adversarial attack effects can also be achieved by using randomization. For instance, Xie \textit{et al.} proposes to first resize the input to random size \shortcite{xie2017mitigating}. After that, a random padding operation is performed to pad zeros around the resized image. Though it may seem simple, this method can significantly improve the robustness of DNN models against adversarial attack.

\subsubsection{Input Transformation} Another type of popular defense methods is input transformation. Its key idea is to perform various transformations, such as bit-depth reduction, lossy compression and variance minimization on adversarial examples to mitigate the attack effects \cite{guo2017countering,xu2017feature}. The reported experimental results show that these methods can achieve balance between robustness against attack and computation overhead.

\section{Motivations}
Despite the abundance of researches on adversarial attack methods described in Section 2, existing approaches still suffer from several inherent drawbacks--in particular from the perspective of practical deployment.

\textbf{Long Generation Time} Iteration-based approaches predominate among current state-of-the-art methods, including PGD and C\&W. Consequently, generating adversarial examples using iteration is time-expensive and requires extensive computational resources, especially for the targeted attack. For example, to achieve a high ASR, C\&W method takes hours to generate 100 large-size adversarial examples on a GPU. Such long generation time makes launching the adversarial attack in real-time setting infeasible.

\textbf{High Memory Cost} Using an iteration-free generative model-based attack promises to avoid long generation time \cite{xiao2018generating,poursaeed2018generative,baluja2017adversarial}. However, in these existing works if the attackers wants to achieve targeted attack to a specific class, they have to train different generator models for different targeted classes. For example, to prepare for the targeted attack to 1000 classes in the ImageNet dataset, in total 1000 different generator models have to be trained and stored, thereby causing massive memory cost.

\textbf{Insufficient Robustness} To date, most adversarial example generation is based on the search over the entire input size instead of focusing on the critical part of legitimate object content. Noticing this phenomenon, many defense methods have been developed to improve defense performance via integrating this information into the defense scheme. For instance, Luo \textit{et al.} propose to mask out the background regions with little transformation performed on the critical areas \cite{luo2015foveation}. Similarly, Prakash \textit{et al.} propose to use pixel deflection  to denoise and reconstruct the input by locally redistributing pixels under the guidance of the object position \shortcite{prakash2018deflecting}. Consequently, such well-designed defense schemes make the existing adversarial attack exhibit insufficient robustness.

\textbf{Low Transferability} Most adversarial attack methods can achieve high ASR in the white-box attack scenario. However, in real-world applications, black-box attack is a more common environment setting. In such cases, the transferability of the generated adversarial examples is important to ensure a successful attack. However, to date on large-scale datasets and large DNN models, the existing adversarial attack approaches exhibit low transferability, thereby impeding the feasibility of launching real-life black-box attack. 

\textbf{Our Motivation} Motivated to redress the above challenges plaguing the existing adversarial attack methods, we aim to develop an adversarial attack method that can 1) generate each adversarial example in a real-time manner; 2) require only one model for different targeted classes; 3) exhibit strong robustness against the-state-of-the-art defense techniques; and 4) exhibit high transferability in the black-box attack scenario. To fulfill those requirements, we develop CAG, an attack method with fast generation speed, low memory cost, improved robustness and high transferability. Next, we describe the model training and attack generation schemes of CAG in detail.

\section{CAG: Content-aware Adversarial Attack Generator}

\subsection{Overall Architecture}
\begin{figure}[ht]
\centering
\includegraphics[width=0.45\textwidth]{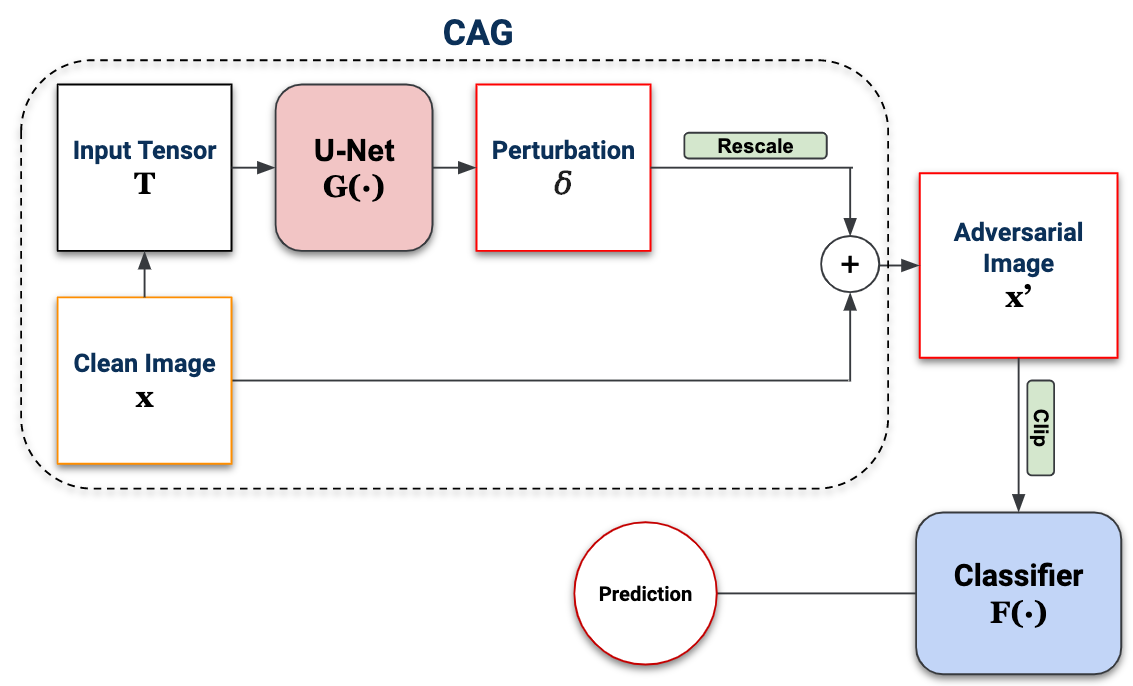}
\caption{Overall architecture of CAG.}
\label{overall}
\end{figure}
Figure \ref{overall} illustrates the overall architecture of CAG. To generate an adversarial image, an input tensor $T$ is first constructed based on the given clean image $x$, true label $i$ and targeted label $t$. Then a generator model $G(\cdot)$, in the format of U-Net, is used to generate the perturbations $\delta$ from $T$. After that, $\delta$ is scaled to a fixed $L_2$ norm and added to $x$. Finally, after clipping out-of-range values, the adversarial image $x'$ is ready to mislead the classifier from original true class $i$ to the targeted class $t$. 

\textbf{Fast Generation Speed using U-Net} CAG utilizes U-Net as the underlying generative model. Therefore, when compared with other iteration-based attack methods, U-Net-based approach avoids time-expensive iterative procedure, and hence makes real-time generation of adversarial examples possible. 

\subsection{Building Input Tensors}
\begin{figure}[ht]
\centering
\includegraphics[width=0.45\textwidth]{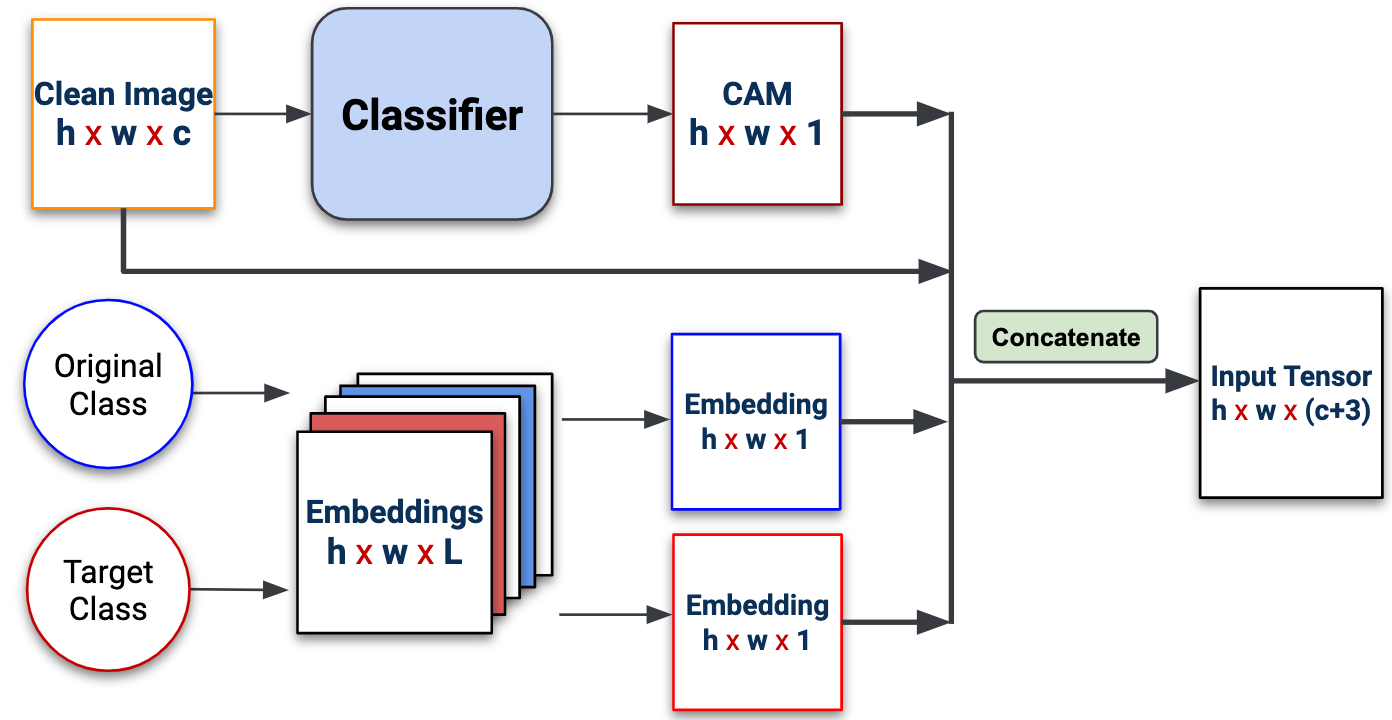}
\caption{Building input tensor for CAG. The target class is randomly selected and given in training and testing phase, respectively.}
\label{input_tensor}
\end{figure}

\textbf{Single Generative Model via Label Embedding} As mentioned in Section 3, the main drawback of generative model-based attack is a need for massive amount of models for different target classes. To address this problem, we encode the class label information into the input tensor $T$ for U-Net. Figure \ref{input_tensor} shows the overall procedure of constructing $T$. Here the dimension of the clean image is denoted as $h \times w \times c$ where $h, w, c$ represent height, width and number of channels, respectively. Then during training phase, an embedding layer $E_{hwL}$ with the size of $h \times w \times L$, where $L$ is the number of valid classes, is trained to encode the label information. Specifically, in the forward propagation pass, a targeted class $t$ is randomly selected for each training data $x_i$. The target label $t$, as well as the true label $i$, are used to extract the corresponding slices $E_{t}$ and $E_{i}$ from $E_{hwL}$, where $E_{k}=E_{::k}$ denotes the k-th front slice of the embedding layer $E_{hwL}$. Then in the backward propagation phase, $E_{t}$ and $E_{i}$ are updated to help $E_{hwL}$ capture more class information for this training data. After being trained on the entire dataset, the embedding layer $E_{hwL}$ learns the important class encoding information and thereby ensuring only one U-Net model is sufficient for different target classes.

\textbf{Enhanced Attack Robustness using CAM} Besides using an embedding layer, the construction of input tensor $T$ also utilizes the information of CAMs. Classifiers make decisions based heavily on the hot areas of CAM because they contain the most discriminative information of an image. Therefore, defense methods cannot make substantial modifications in these critical area, otherwise they can easily cause misclassifications. Taking advantage of this behavior, we place the perturbations only on the hot areas of CAM to enhance the robustness of our attack against many defense schemes. To achieve this increment in robustness, the position of the object in the image needs to be integrated into the input tensor, which can be reflected by the CAM. As shown in Figure \ref{input_tensor}, another component of input tensor $T$ is $CAM_{x}^{i}$ with the size of $h \times w \times 1$, which is the CAM with respect to input $x_{i}$ and its true label $y_{i}$. Consequently, we denote $\tau$ as the concatenating operation, and the final input tensor is constructed as follows:
\begin{equation*}
    T=\tau(x_{i}, E_{i}, E_{t}, CAM_{x}^{i}),
\end{equation*}
where the size of $T$ is $h \times w \times (c+3)$.

\subsection{Training CAG}

\begin{algorithm}
\SetAlgoLined
\textbf{Input:} {dataset $X=\{x_1,...,x_m\}$, true labels $y \in \{1,...L\}$, classifier $F(\cdot)$, input tensor $T$, desired perturbation $L_2$ norm.}\\
\textbf{Result:} {trained CAG $G(\cdot)$, embedding layer $E_{hwL}$.}\\
Random initialize $G(\cdot)$, E$_{hwL}$.\\
 \For{$x_{i}$, $y_{i}$ \textbf{in} dataset}{
  $y_{t}$ = get\_random\_target($y_{i}$), $t\neq i$\;
  $E_{i}$ = ($E_{::i}$), $E_{t}$ = ($E_{::t}$); $h \times w \times 1$\;
  $CAM_x^i$ = cam\_generator($x$, $y_{i}$)\;
  $T$ = concat($x_{i}$, $E_{i}$, $E_{t}$, $CAM_x^i$)\;
  $\delta$ = drop\_out ($G(T)$)\;
  $\delta$ = L2\_norm\_adjust($\delta$)\;
  adversarial\_img = $x_{i}'$ = clip($x_{i}$ + $\delta$) \;
  $y_{pred}$ = $F(x_{i}')$\;
  $CAM_{x'}^{t}$ = cam\_generator($x'$, $y_{t}$)\;
  \textbf{Loss} = CrossEntropy($y_{t}, y_{pred}$) + $\beta\cdot$$\Vert CAM_{x}^{i} - CAM_{x'}^{t}\Vert^2$\;
  update($G(\cdot)$); update($E_{hwL}$)\;}
\caption{CAG Training Algorithm}
\label{al}
\end{algorithm}

Next, we describe the details of CAG training procedure. Our objective is to get $G(\cdot)$ and $E_{hwL}$ to achieve:
\begin{equation*}
    F(clip(G(T)+ x_{i}))=y_{t}.
\end{equation*}
In this scenario, the embedding layer $E_{hwL}$ is treated as a model parameter that can be learned, so that $y_{t}$ can be any selected label from $\{1, ..., L\}$. Therefore, we can formulate an effective loss function $Loss$, and use existing optimization algorithms to perform training as follows.

First, in order to keep the perturbations imperceptible, we scale the perturbations using the $L_2$ distance metric. In other words, we keep all the perturbations at a fixed $L_{2}$ norm to constrain the attack strength of the noise in a fixed amount.

Then we feed the generated adversarial example $x_{i}'$ to the classifier $F(x_{i}')$ to produce the prediction $y_{pred}$. We define $Loss_{target}$ as the cross-entropy with respect to the one-hot label of the targeted class. Therefore, to ensure the generated adversarial examples can fool the classifier, $Loss_{target}$ is formulated as:
\begin{equation*}
    Loss_{target} = CrossEntropy(y_{t}, y_{pred}).
\end{equation*}

Meanwhile, the CAM of the targeted class $t$ for $x'$ is computed and denoted as $CAM_{x'}^{t}$. We aim to concentrate the adversarial noise on the critical areas which contain the legitimate object content, so that the $CAM_{x'}^{t}$ for the adversarial examples would not be significantly changed compared to $CAM_{x}^{i}$. In other words, to satisfy the similarity between $CAM_{x'}^{t}$ and $CAM_{x}^{i}$, we need to minimize the their $L_2$ distance. Therefore, $Loss_{CAM}$ is defined to lead the distribution of the noise:
\begin{equation*}
    Loss_{CAM} = \Vert CAM_{x}^{i} - CAM_{x'}^{t}\Vert^2.
\end{equation*}

Finally, the new loss function is formulated as:
\begin{equation*}
Loss =  Loss_{target} + \beta \cdot  Loss_{CAM},
\end{equation*}
where $\beta$ controls the magnitude of $Loss_{CAM}$. We then iteratively optimize the CAG as well as $E_{hwL}$ by minimizing the $Loss$ function. The details of our approach to train the CAG are summarized in Algorithm \ref{al}. 

\textbf{Improve Transferability via Noise Dropout} It is worth noting that before directly adding the noise on $x_{i}$, we propose to apply a dropout layer with probability $p$ in the training phase. As a result, dropout layer can eliminate over-fitting problem to the current classifier and achieve better performance in black-box attack scenario by increasing the transferability. The extensive experimental results are given in the next section.

\begin{figure}[ht]
\centering
\includegraphics[width=0.45\textwidth]{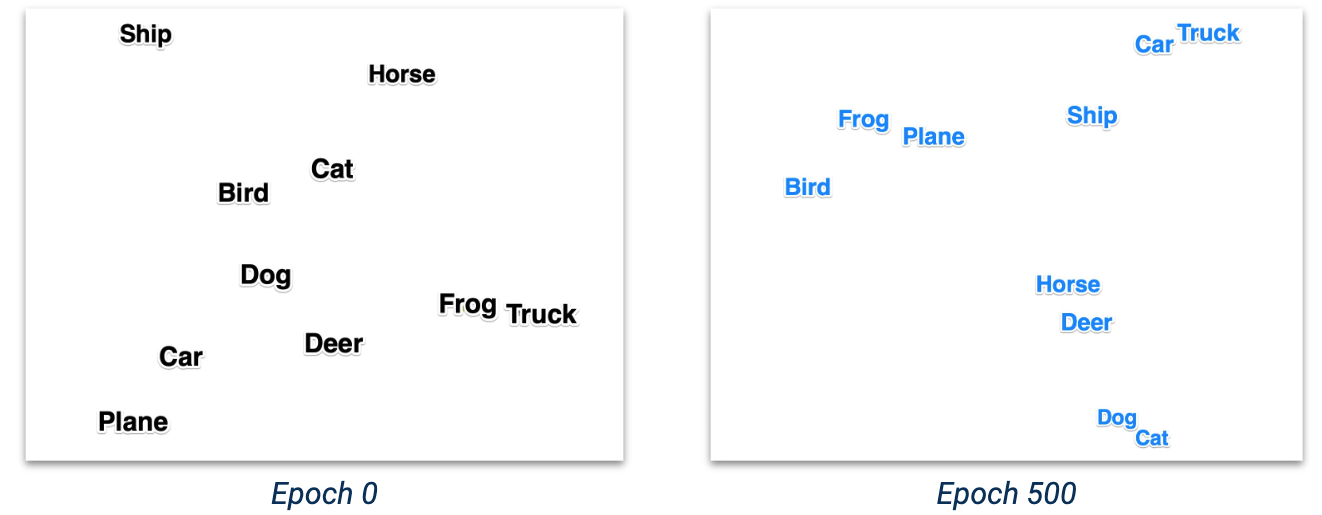}
\caption{Embeddings of different classes after using T-SNE to reduce the dimension (CIFAR-10).}
\label{embedding}
\end{figure} 

\begin{figure}[ht]
\centering
\includegraphics[width=0.45\textwidth]{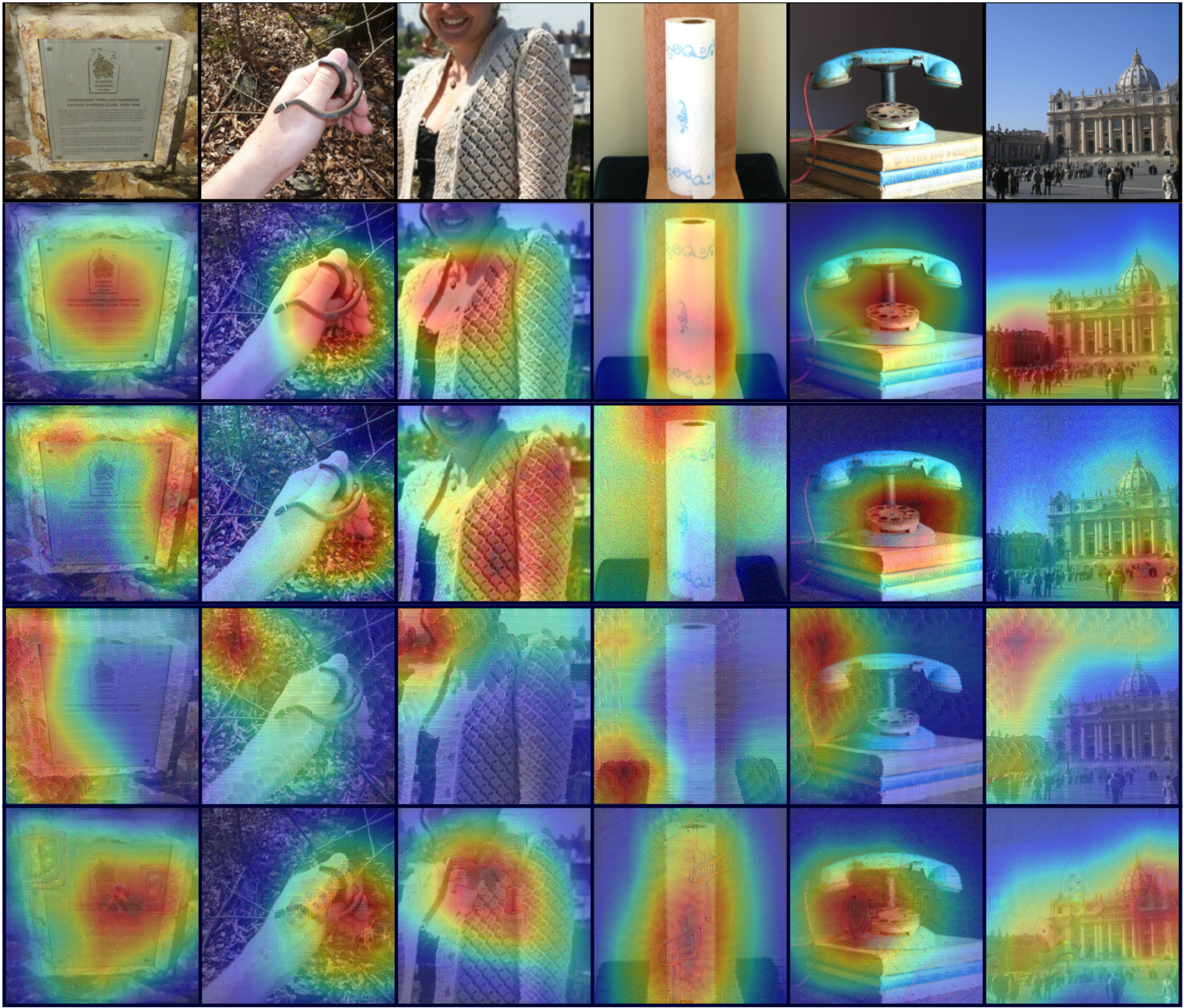}
\caption{CAM attention visualization. From top row to bottom row: clean images, CAM of clean images, CAM of PGD's adversarial images, CAM of GAP's adversarial images, CAM of CAG's adversarial images.}
\label{cams_compare}
\end{figure}
After preparing CAG and $E_{hwL}$ to perform attacks, we visualize the $E_{hwL}$ to demonstrate the effectiveness of this embedding layer. We show the examples using CIFAR-10 dataset, thus the size of the embedding layer is $32\times32\times10$ for 10 classes \cite{krizhevsky2009learning}.
For better visualization, T-SNE is applied to reduce the each class embedding's dimension to 2 \cite{maaten2008visualizing}. As we can see from Figure \ref{embedding}, at epoch 0, class embeddings are initialized and distributed randomly. However, at epoch 500, embeddings of similar classes are close to each other, such as car-truck, horse-deer, and dog-cat. 
Therefore, the local distance between similar classes suggests that our approach creates a useful set of embeddings.

We also show the attention regions using CAM for adversarial examples generated by different attack methods. As shown in Figure \ref{cams_compare}, compared with the clean images, I-FGSM and GAP achieve targeted attack by misleading the network's attention. However, we believe that changing the attention would make adversarial images vulnerable to designed defense mechanisms. Interestingly, as can be seen in the last row of Figure \ref{cams_compare}, the adversarial images generated by CAG do not suffer from this problem. Malicious perturbations are constrained to locate in the discriminative areas, so that CAG's adversarial examples are robust enough circumvent detection and defense methods.

\section{Experimental Results}

\subsection{Experiment Design}
To evaluate the effectiveness of CAG, we conduct extensive experiments on CIFAR-10 \cite{krizhevsky2009learning} and ImageNet \cite{deng2009imagenet} dataset. 
We perform white-box and black-box attacks by using a pool of 6 different classifiers: 
ResNet-18 (RN-18), ResNet-34 (RN-34); VGG-11, VGG-13; DenseNet-121 (DN-121), DenseNet-169 (DN-169) \cite{he2016deep,simonyan2014very,huang2017densely}. The top-1 classification accuracy is above $92\%$ (CIFAR-10) and above $70\%$ (ImageNet) for all classifiers. We use ResNet-18 to generate CAMs for all experiments. 
We set $\beta=3$ for both datasets because higher $\beta$ enforces too much restriction and can reduce ASR. Then we train CAG using SGD with Nesterov momentum. The initial learning rate is set to $5e^{-2}$ and gradually decayed to $1e^{-6}$ using a cosine annealing curve. During training, a target label is randomly picked from all incorrect classes for each data point. On CIFAR-10, the CAG is trained for a total of $500$ epochs using the batch size of $256$. On ImagetNet, we train the CAG for $20$ epochs with batch size of $64$.
The $L_{2}$ norm of adversarial perturbations is set to $0.1$ for both datasets.

We compare our proposed method with other existing attack algorithms: I-FGSM, PGD, and C\&W. We use FoolBox in PyTorch \cite{rauber2017foolbox} to generate these adversarial examples.
Our experiments are performed on NVIDIA Tesla V100 GPU. 

\subsection{CIFAR-10}
\begin{table}[ht]
\centering
\scalebox{0.95}{
\begin{tabular}{|l|c|c|c|c|}
\hline
\textit{} & \textbf{ASR} & \textbf{Acc.} & \textbf{$L_2$} & \textbf{Time} \\ \hline
I-FGSM & 99.53\% & 0.05\% & 0.106 & 13m24s \\ \hline
PGD & 99.56\% & 0.22\% & 0.106 & 12m56s \\ \hline
C\&W & \textbf{99.85}\% & \textbf{0.14}\%& \textbf{0.009} & $>$10h \\ \hline
CAG & 97.29\% & 1.4\% & 0.100 & \textbf{1.44s} \\ \hline
\end{tabular}}
\caption{Comparison of adversarial examples generated by CAG and other methods on ResNet-18 (CIFAR-10).}
\label{table1}
\end{table}

\begin{table*}[h]
\centering
\scalebox{0.90}{
\begin{tabular}{|l|c|c|c|c|c||c|}
\hline
\textit{\textbf{Attacks / Classifiers}} & RN-34 & VGG-11 & VGG-13 & DN-121 & DN-169 & \textbf{Average} \\ \hline
I-FGSM & 30.45\% & 21.39\% & 23.88\% & 29.01\% & 28.70\% & 26.69\% \\ \hline
PGD & 45.38\% & 28.46\% & 34.90\% & 41.15\% & 40.05\% & 37.99\% \\ \hline
C\&W & 7.57\% & 8.88\% & 8.40\% & 6.61\% & 7.70\% & 7.85\% \\ \hline
CAG $p=0.0$ & 86.47\% & 66.46\% & 94.09\% & 83.64\% & 85.74\% & 83.78\% \\ \hline
CAG $p=0.1$ & 89.02\% & 70.51\% & \textbf{95.10\%} & 87.01\% & 88.52\% & 85.92\% \\ \hline
CAG $p=0.2$ & 90.83\% & 74.31\% & 94.93\% & 88.93\% & 90.37\% & 87.85\% \\ \hline
CAG $p=0.3$ & \textbf{91.83\%} & \textbf{77.81\%} & 94.89\% & \textbf{90.49\%} & \textbf{91.31\%} & \textbf{89.24\%} \\ \hline
\end{tabular}}
\caption{Comparison of transferbility (ASR) of various attack methods and CAG with different dropout rate $p$ on ResNet-18 in black-box scenario (CIFAR-10).}
\label{table2}
\end{table*}

\begin{table*}[ht!]
\centering
\scalebox{0.90}{
\begin{tabular}{|l|l|c|c|c|c|c|c|c|c|}
\hline
\textbf{} & \textbf{Storage} & \multicolumn{4}{c|}{\textbf{White-box}} & \multicolumn{4}{c|}{\textbf{Black-box}} \\ \hline
\textit{\textbf{Attacks / Classifiers}} &  & RN-18 & VGG-11 & DN-121 & \textbf{Average} & RN-34 & VGG-13 & DN-169 & \textbf{Average} \\ \hline
GAP Unet (5T) & 30 MB $\times$ 5 & 97.98\% & \textbf{98.45\%} & \textbf{97.85\%} & \textbf{98.09\%} & 82.97\% & 85.69\% & 88.31\% & 85.66\% \\ \hline
GAP ResNet (5T) & 30 MB $\times$ 5  & 91.02\% & 94.25\% & 90.58\% & 91.95\% & 76.40\% & 86.27\% & 78.33\% & 80.33\% \\ \hline
GAP (1000T) & 30 MB $\times$ 1000  & N/A & N/A & N/A & N/A & N/A & N/A & N/A & N/A \\ \hline
CAG (5T) & 222 MB & \textbf{98.52\%} & 97.71\% & 96.91\% & 97.71\% & \textbf{95.45\%} & \textbf{94.34\%} & \textbf{94.06\%} & \textbf{94.62\%} \\ \hline
CAG (1000T) & 222 MB & 97.79\% & 97.01\% & 96.62\% & 97.14\% & 93.38\% & 94.28\% & 92.61\% & 93.42\% \\ \hline
\end{tabular}}
\caption{Storage and ASR comparison of adversarial examples generated by CAG and GAP (ImageNet). Both are trained on ensemble of models: RN-18, VGG-11 and DN-121. 5T and 1000T represents 5 and 1000 targeted classes, respectively. Due to the limitation of storage and impractical training time, we can not report the attack results on GAP with 1000T.}
\label{table5}
\end{table*}

\begin{table*}[ht!]
\centering
\scalebox{.90}{
\begin{tabular}{|l|c|c|c|c|c|c|c|c|c|c|}
\hline
\multicolumn{1}{|c|}{} & \multicolumn{5}{c|}{\textbf{White-box}} & \multicolumn{5}{c|}{\textbf{Black-box}} \\ \hline
\textit{\textbf{Defense Methods}} & RN-18 & VGG-11 & DN-121 & \textbf{Average} & \textbf{\begin{tabular}[c]{@{}c@{}}Average\\ I-FGSM\end{tabular}} & RN-34 & VGG-13 & DN-169 & \textbf{Average} & \textbf{\begin{tabular}[c]{@{}c@{}}Average\\ I-FGSM\end{tabular}} \\ \hline
None & 0.59\% & 1.03\% & 1.50\% & \textbf{1.04\%} & 4.58\% & 2.10\% & 1.51\% & 3.85\% & \textbf{2.49\%} & 42.17\% \\ \hline
Pixel Deflection & 24.75\% & 26.39\% & 31.76\% & 27.63\% & \textbf{14.76\%} & 33.83\% & 23.44\% & 43.45\% & \textbf{33.57\%} & 57.25\% \\ \hline
Randomization & 3.06\% & 3.07\% & 6.80\% & \textbf{4.31\%} & 22.19\% & 5.26\% & 2.77\% & 10.23\% & \textbf{6.09\%} & 43.90\% \\ \hline
Bit Depth Reduction & 4.80\% & 7.33\% & 10.41\% & \textbf{7.51\%} & 10.22\% & 11.12\% & 10.18\% & 17.85\% & \textbf{12.96\%} & 50.08\% \\ \hline
JPEG Compression & 5.81\% & 7.63\% & 11.05\% & \textbf{8.16\%} & 12.62\% & 12.17\% & 10.48\% & 17.78\% & \textbf{13.48\%} & 49.36\% \\ \hline
\end{tabular}}
\caption{Classification accuracy of CAG's adversarial images versus I-FGSM's after applying defense mechanisms (ImageNet).}
\label{table6}
\end{table*}

We first evaluate our proposed CAG on CIFAR-10 in white-box scenario. The classifier is set to be ResNet-18, and the classification accuracy on clean images achieves 93.48\% for 10,000 validation images. To evaluate the targeted attack algorithms, ASR is used as the performance metric. 

\subsubsection{Low Computation Time} We generate 10,000 adversarial examples in CIFAR-10 validation set, and each image is targeted to a randomly incorrect class. The ASR can reach $97.29\%$ on the ResNet-18. We compare our proposed CAG with other state-of-art targeted attack methods. Similar to the procedure we use to evaluate CAG, we also choose attack targets in random manner. As for C\&W, we only report first 1000 images targeted on random classes. Since the $L_{2}$ norm for CAG is set to be $0.1$, for fair comparison, we try to keep $L_{2}$ norm around similar range for I-FGSM and PGD. Therefore, $\epsilon$ and $\alpha$ is set to $0.1$ and $0.035$, respectively. The maximum iteration is set to 50. When using C\&W attack, we perform 10 iterations of binary search and run 10,000 iterations of gradient descent with learning rate at $0.005$ using the Adam optimizer. We only generate 1,000 images using C\&W attack. 
As can be seen from the Table \ref{table1}, our attack achieves comparable results compared with I-FGSM, PGD, and C\&W. However, our attack has much lower inference time of only 1.44 seconds compared of 12 minutes 56 seconds of PGD and more than 10 hours of C\&W attack--a more than 500-fold speedup. The ability to generate a large number of adversarial images in a such a small time makes our attack method practical in real-time applications.

\subsubsection{High Transferability} CAG always has ASR greater than $95\%$ in white-box attack scenario. However, considering black-box attack, when attackers have no access to architecture and parameters of the classifier, ASR is not as high as the white-box scenario. To address this high transferbility requirement, we propose to drop out part of the perturbation before adding it on the benign image during training phase. As a result, CAG generalizes better and is less prone to over-fitting to a particular classifier. Hence, the transferability of the adversarial examples to new classifiers increases. We train 4 CAG models using ResNet-18 with dropout probability $p=0.0$, $p=0.1$, $p=0.2$ and $p=0.3$. The ASR for 10,000 validation images (only 1000 images for C\&W) targeted on random incorrect classes are reported. Table \ref{table2} reveals that even without dropout, CAG still has better performance in black-box results compared with other methods. Furthermore, the transferability of adversarial examples improves with increasing dropout probability.

\subsection{ImageNet}

We also evaluate the CAG on ImageNet. In our experiments, CAG takes a long time to converge when trained with a single classifier. Thus to accelerate the training process and perform stronger attack, we train CAG with an ensemble of ResNet-18, VGG-11 and DenseNet-121. When training with an ensemble of classifiers, we observe that the CAG does not suffer from over-fitting as much as training with only one classifier. Hence, unlike the best configuration in CIFAR-10 where $p=0.3$, we reduce the perturbation dropout to $p=0.1$ in this case.

To explicitly demonstrate the performance of our proposed method, we compare our results with GAP \cite{poursaeed2018generative}. To create a fair comparison, we implement GAP with two architectures and keep the configuration the same as our method. The first GAP uses identical generative architecture to ours, so we denote it GAP U-Net. The second GAP has the same architecture used in GAP's original paper, which we denote it GAP ResNet. However, to perform targeted attack, GAP requires 1 model for each targeted class. Because we do not have enough resources to train 1,000 GAP models to have a comprehensive evaluation, we train 5 models for each architecture targeted at these following random chosen classes: \textit{black swan, Tibetan terrier, tiger beetle, cliff dwelling, hook.} 

\subsubsection{Low Memory Cost} The comparison result is shown in Table \ref{table5}. 10,000 benign images are randomly picked from the validation dataset to do the evaluation. We use CAG to generate adversarial examples targeted at the same 5 selected labels for fair comparison. In addition, since our proposed CAG can perform the comprehensive targeted attack on all 1,000 classes, we also generate adversarial images crossing all classes. In the table, 5T means ASR are evaluated on a pool of the same 5 targeted classes using 10,000 images in ImageNet evaluation dataset. In the last row, 1000T means that 10,000 images are targeted to any randomly selected label from all 1,000 classes. As can be seen from the table, to perform comprehensive attacks to all 1000 classes of ImageNet, our model takes 222MB of storage: 30MB for model's weights, and 192MB for the embeddings. However, other generative models can take up to 30MB $\times$ 1000 $\approx$ 30GB for storage to attack all classes.
Moreover, as shown in Table \ref{table5}, for seen classifiers, ASR is above 90\% for all approaches. On the one hand, while targeting 5 selected classes, adversarial images generated by GAP U-Net and CAG have comparable performance. On the other hand, by analyzing the result of unseen classifiers, we can see that CAG outperforms GAP. ASR of CAG can reach to 93.42\% for 1,000 target labels in black-box scenario. To sum up, our proposed CAG is more practical to perform general targeted attack while keeping high ASR and transferability.

\subsection{Breaking Defenses}

\subsubsection{Enhanced Robustness} Finally, we study the robustness of adversarial examples generated by CAG on ImageNet. We prepare CAG trained on the ensemble of ResNet-18, VGG-11 and DenseNet-121. Using the optimal setting, the dropout probability is set to $0.1$. Since it is meaningless to protect images that are originally mis-classified, we evaluate 10,000 (ImageNet) images that are correctly classified by all three classifiers. We use the following configurations:

\subsubsection{Pixel Deflection}
To achieve the strongest defense performance we provide the CAMs of the true class of correctly classified images to guide the pixel deflection (unlike using CAMs of top-5 predictions as the original paper suggests). We set the parameters following the original paper with window=10, deflections=100.
\subsubsection{Randomization}
To perform this defense with optimal parameters, we keep the scale ratio the same as the ratio reported in the original paper. Thus the image size is modified from $224\times224\times3 $ to  $253\times253\times3 $ in our implementation.
\subsubsection{Bit-depth Reduction}
In our experimental setting, we reduce images to 3 bits as the original paper \cite{xie2017mitigating}.
\subsubsection{JPEG Compression}
We perform JPEG compression at quality level 75 out of 100.

Classification accuracy after applying defense methods is shown in Table \ref{table6}. As a result of using CAM guidance in proposed CAG, our attack is robust against defense methods that aim to modify the non-discriminative regions such as pixel deflection and randomization. After using pixel deflection, classifiers accuracy on CAG's generated adversarial images is still low at 27.63\% (white-box) and 33.34\% (black-box). In addition, CAG's adversarial images can bypass the defense effects of input transformation. Bit depth reduction and JPEG compression can not improve the accuracy more than 10\% for white-box and 14\% for black-box setting. Compared with I-FGSM, our attack achieves lower classification accuracy in almost all categories. To sum up, our attack is robust against many defense mechanisms.

\section{Conclusions}
In this work, we propose a generative model to perform targeted adversarial attacks called CAG. With the help of the trainable embedding layer, the supervision of CAMs and random dropout, CAG is able to produce robust adversarial examples with state-of-art attacking performance and high transferability, while still maintaining low computation time and low memory cost. CAG has many desirable properties of an adversarial attack method, and therefore outperforms many other methods and can launch a real-time robust attack against many modern DNN systems.

\section*{Acknowledgments}
Partial financial supports by AFRL (USA) under Grant No. FA8750‐18‐2‐0058.
\clearpage

\bibliography{bibliography.bib}
\bibliographystyle{aaai.bst}

\end{document}